# Rain Removal via Shrinkage-Based Sparse Coding and Learned Rain Dictionary

Chang-Hwan Son and Xiao-Ping Zhang

**Abstract— This paper introduces a new rain removal model based on the shrinkage of the sparse codes for a single image. Recently, dictionary learning and sparse coding have been widely used for image restoration problems. These methods can also be applied to the rain removal by learning two types of rain and non-rain dictionaries and forcing the sparse codes of the rain dictionary to be zero vectors. However, this approach can generate unwanted edge artifacts and detail loss in the non-rain regions. Based on this observation, a new approach for shrinking the sparse codes is presented in this paper. To effectively shrink the sparse codes in the rain and non-rain regions, an error map between the input rain image and the reconstructed rain image is generated by using the learned rain dictionary. Based on this error map, both the sparse codes of rain and non-rain dictionaries are used jointly to represent the image structures of objects and avoid the edge artifacts in the non-rain regions. In the rain regions, the correlation matrix between the rain and non-rain dictionaries is calculated. Then, the sparse codes corresponding to the highly correlated signal-atoms in the rain and non-rain dictionaries are shrunk jointly to improve the removal of the rain structures. The experimental results show that the proposed shrinkage-based sparse coding can preserve image structures and avoid the edge artifacts in the non-rain regions, and it can remove the rain structures in the rain regions. Also, visual quality evaluation confirms that the proposed method outperforms the conventional texture and rain removal methods.**

## I. INTRODUCTION

Rain forms structures on captured images. This means that rain structures can prevent computer vision algorithms (e.g., face/car/sign detections, visual saliency, scene parsing, etc.) from working effectively [1]. Most computer vision algorithms depend on feature descriptors such as scale invariant feature transform (SIFT) [2] and histogram of oriented gradients (HOG) [3]. These descriptors are designed based on the gradient's magnitude and orientation, and thus the rain structures can have negative effects on the feature extractor. For this reason, rain removal is a necessary tool. Rain removal can preserve the details of objects and suppress the rain structures and thus, it can be used to detect the visual saliency [4] on the rain images. Moreover, recently, self-driving [5] is a hot problem in the vehicle industry. To realize this, bad weather conditions [6] including rain, snow, or haze should be considered. Therefore, low-level computer vision algorithms, such as rain and haze removals, are essential in smart cars. Application of rain removal is not limited only to computer vision problems in bad weather conditions. In general, rain removal first detects certain types of image structures, and then removes the detected structures from input images. Therefore, rain removal approaches can be applied to similar problems that appear in computer graphics and image processing. For example, rain removal approaches can used for texture [7], stripes [8], waterfall effects [9], or other types of artifacts [10] removal.

### A. Related Works

In most cases, rain structures can be described by vertical and diagonal edges. However, in some cases, rain structures appear with other types of patterns. Given an input rain image, several approaches can be considered to remove the rain structures. The first approach is to directly use the conventional texture removal algorithms. The morphological component analysis (MCA) [11] and relative total variation (RTV) [7] are sophisticated methods to remove the textures. The MCA algorithm [11] uses parametric-based dictionaries, which indicate the basis vectors of the discrete cosine transform (DCT) and curvelet wavelet transform (CWT). Different types of DCT and CWT dictionaries can discriminate between the textured and non-textured parts. Especially, the CWT can detect the anisotropic structures and smooth curves/edges, while the DCT can represent periodic patterns. However, it has not been proven that the two parametric-based dictionaries are still effective to separate the rain textures from the input rain images. Another RTV-based texture removal [7] can be used for rain removal if the rain structures are fine textures. In [7], the RTV is defined as the absolute value of the sum of the spatial gradients calculated at every local region and it is shown that the RTV is useful to distinguish the rain structures from the main structures (e.g., large edges/lines). However, for images with heavy rain pattern, the RTV model may fail to discriminate between the rain and original main textures, thereby removing the rain and main textures at the same time.

The second approach is to describe the features for the rain structures, and then remove the rain structures from the input rain-patterned images [1]. To detect the rain structures, hand-crafted rain features can be designed based on elliptical

This work was supported by the Natural Sciences and Engineering Research Council of Canada (NSERC) under Grant RGPIN239031.

C.-H. Son is with the Department of Electrical and Computer Engineering, Ryerson University, ON M5B2K3 Canada (E-mail: changhwan76.son@gmail.com.)

X.-P. Zhang is with the Department of Electrical and Computer Engineering, Ryerson University, ON M5B2K3 Canada (E-mail: xzhang@ryerson.ca).

shape [12] or high visibility and low saturation [13]. After detecting the rain regions via a handcrafted feature descriptor, rain structures can be removed via nonlocal filtering or image inpainting. However, recent trend is to adopt representation learning approach [14] (e.g., dictionary learning [15] or deep learning [16]) rather than designing handcrafted features. Representation learning can automatically extract useful features from raw data and it has shown powerful performance for image restoration and image classification problems. For this reason, representation learning has replaced traditional handcrafted design. Following this trend, online dictionary learning has been used to represent rain and non-rain image structures in an input image [1]. In this method, to separate the rain dictionary part from the whole dictionary, handcrafted HOG feature descriptor was used with an assumption that rain structures have vertical and diagonal edges with high variations. However, rain structures are not restricted to only vertical and diagonal edges with high variations. Therefore, the HOG descriptor has limited ability to classify the rain dictionary part from the learned whole dictionary. Even though this method can increase rain removal performance when the rain structures are removed, it may remove object details as well. To overcome this drawback, the depth of field (DOF) can be considered to roughly represent the non-rain regions [17]. However, DOF is also a handcrafted feature. Moreover, the use of DOF is not the main algorithm for the sparse-coding-based rain removal. Rather, it is regarded as a pre-processing step that can be applied to any rain removal algorithms. Actually, other visual saliency algorithms [4] can be used instead of DOF as well. Recently, a more elegant rain removal method was presented based on the discriminative sparse coding [18]. The key idea of this paper is to make the sparse codes of rain and non-rain dictionaries mutually exclusive. However, as discussed in [18], perfect mutual exclusivity cannot be guaranteed for some rain images with similar structures between rain and objects, which leads to unsatisfactory results. In [18], the initial rain dictionary is designed by using motion kernel with the dominant gradient orientation of input rain image. Thus, this initialization can affect final rain removal performance. Rain structures still exist on the rain regions even though object's details are preserved with the learned non-rain dictionary. We will show this in our experimental results. There is another related work [19] that removes dirt and water droplet on the captured images through a window. In this paper, convolutional neural network (CNN) was used to map corrupted image patches onto the clean patches in a supervised manner. However, this method cannot be directly applied to rain removal. In the case of rain images, it is not easy to collect the corrupted and clean patch pairs as it is necessary in a supervised method. Also, rain structures have variety of forms in size and shape, and thus direct mapping from the corrupted patches to the clean ones may not work properly because rain detection and rain removal process are incorporated into the CNN model. However, the learned rain features via the CNN can be used for rain detection in an unsupervised manner, which requires an additional classifier (e.g., logistic regression or support vector machine), and then rain removal process, such as image inpainting or nonlocal filtering can be conducted to remove rain structures.

Meanwhile, there are video de-raining methods that are based on temporal and chromatic priors [20] Gaussian mixture model [21], analysis of rain layers in transformed domain [22], etc. However, this paper focuses on the rain removal for a single image, and thus more details will not be discussed in this paper and the reader may refer to related literatures [20-23].

### B. Motivation

Online dictionary learning requires the handcrafted HOG descriptor [1] to separate the rain dictionary part from whole dictionary. However, the HOG descriptor cannot model various types of rain structures, and thus in this paper, offline dictionary learning is adopted. That is, rain dictionary is learned from the rain training images, not from the input rain image. However, even the rain training images can include non-rain regions. Therefore, in this paper, masked images are generated manually, and then used during the offline dictionary learning to indicate the rain regions. By doing this, the learned rain dictionary can represent various types of rain structures. Also, non-rain dictionary is separately learned from natural training images.

Assuming that the rain and non-rain dictionaries are given, rain removal can be achieved by forcing the sparse codes of the rain dictionary to be zero vectors, as follows:

$$\mathbf{R}_i \mathbf{x} \approx \mathbf{D}\boldsymbol{\alpha}_i = \begin{bmatrix} \mathbf{D}^n \mathbf{D}^r \end{bmatrix} \begin{bmatrix} \boldsymbol{\alpha}_i^n \\ \boldsymbol{\alpha}_i^r = \mathbf{0} \end{bmatrix} \tag{1}$$

where $\mathbf{R}_i$ indicates the operator [24,25] that extracts a patch from input rain image ($\mathbf{x}$) at the $i$th pixel position. If $\mathbf{x}$ is the image vector with the size of $M \times 1$ and the patch size is $m \times m$, the matrix $\mathbf{R}_i$ will be $m^2 \times M$ in size. In this paper, boldface lower case indicates vectors, whereas boldface uppercase indicates matrices. $\mathbf{D} \in \mathfrak{R}^{m^2 \times 2K}$ is the dictionary set in which each column vector corresponds to the basis vector which is also called signal-atom. The dictionary set is composed of $\mathbf{D}^n \in \mathfrak{R}^{m^2 \times K}$ and $\mathbf{D}^r \in \mathfrak{R}^{m^2 \times K}$. These are the learned non-rain and rain dictionaries, respectively. In (1), the left term shows that the extracted input rain patch $\mathbf{R}_i \mathbf{x} \in \mathfrak{R}^{m^2 \times 1}$ can be approximated by the linear combination of the dictionary set $\mathbf{D}$ and the corresponding sparse code $\boldsymbol{\alpha}_i \in \mathfrak{R}^{2K \times 1}$. Any kinds of sparse coding algorithms [15] can be used to estimate the $\boldsymbol{\alpha}_i$ from the rain patch ($\mathbf{R}_i \mathbf{x}$). The estimated sparse code $\boldsymbol{\alpha}_i$ can be divided into $\boldsymbol{\alpha}_i^n$ and $\boldsymbol{\alpha}_i^r$ corresponding to the $\mathbf{D}^n$ and $\mathbf{D}^r$, respectively. Here, the sparse code $\boldsymbol{\alpha}_i^r$ is used only to represent the rain structures. Therefore, the rain structures can be removed by assigning the zero vector to the sparse code $\boldsymbol{\alpha}^r$, thereby producing the rain-removed patch, which is represented by $\mathbf{D}^n \boldsymbol{\alpha}_i^n$. Ideally, the rain dictionary $\mathbf{D}^r$ should be used only to reconstruct the rain structures. However, the rain dictionary $\mathbf{D}^r$ can reconstruct the non-rain structures. Since the signal-atoms of the rain and non-rain dictionaries can have some correlations, similar signal atoms in the rain and non-rain dictionaries can be selected via a matching pursuit algorithm. Therefore, for the non-rain patches, the rain removal approach that forces the sparse code $\boldsymbol{\alpha}_i^r$ to be zero vector, can lead to the edge artifacts and detail loss of objects. Fig.1 shows an example of the edge artifacts and contrast loss that appear around the man's shoulder and face regions including eye, lip, and hair. This observation reveals that the shrinkage of the sparse code $\boldsymbol{\alpha}_i^r$ in the non-rain regions is needed. Also, it is expected that this shrinkage approach should be applied to the sparse code $\boldsymbol{\alpha}_i^n$ in the rain regions. Therefore, the main contribution of this paper is to show a new method of shrinking the sparse codes $\boldsymbol{\alpha}_i^r$ and $\boldsymbol{\alpha}_i^n$ for rain removal. We call it shrinkage-based sparse coding method hereafter.

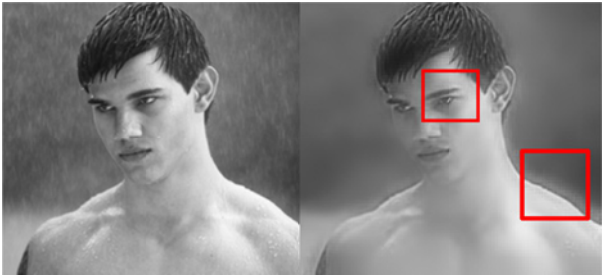

Fig.1. Input rain image (left image) and rain-removed image with edge artifacts and contrast loss (right image).

### C. Proposed Shrinkage-based Sparse Coding vs. Discriminative Sparse Coding for Rain Removal

Before moving to the next section, we first need to discuss more details about the proposed shrinkage-based sparse coding and the conventional discriminative sparse coding [18]. As briefly mentioned in the Introduction, the conventional discriminative sparse coding [18] focuses on generating highly discriminative sparse codes in an iterative manner, whereas our shrinkage-based sparse coding focuses on how to shrink the sparse codes of the fixed rain and non-rain dictionaries to preserve details of objects and avoid edge artifacts. As already pointed out in [18], perfect mutual exclusivity between rain and object structures cannot be ensured. This reveals that the sparse codes of the rain and non-rain dictionaries can be used together to represent object details and rain structures. Even though the discriminative sparse coding is used, similar problem, as shown in Fig. 1, can occur. In other words, rain structures still remain in rain regions or detail may be lost in object regions. Actually, the rain and non-rain dictionaries used in this paper have small correlation about 0.1, meaning that the learned dictionaries have discrimination power to some extent. This is possible because masked rain images are used to extract rain structures from the rain images to learn the rain dictionary. This indicates that even the discriminative sparse coding can have the same problem. Different from initial purpose of the discriminative sparse coding [26] used for image classification, rain removal methods require detection and removal. Therefore, after detecting rain regions via the discriminative sparse coding, it should be considered how to remove the detected rain structures from the rain images, especially for the regions with similar image structures between rain and image layers. This leads to the necessity of the proposed shrinkage-based sparse coding.

There are major differences between the proposed shrinkage-based sparse coding and the conventional sparse coding methods [1,18]. First, in the proposed method, the rain dictionary is learned from the masked rain images that have information regarding which pixel positions contain rain structures. Therefore, the learned rain dictionary can represent various types of real-life rain structures and it can detect rain regions from input rain images directly. In contrast, conventional sparse coding methods [1,18] require a prior knowledge about rain structures. Therefore, the performance of the conventional sparse coding methods depends on that prior knowledge. For example, in the sparse coding method [1], it is assumed that rain structures can be represented by vertical and diagonal edges with high variations. Based on this prior knowledge, the handcrafted HOG descriptor is used to separate the rain dictionary from the whole dictionary. For long and thick rain structures, the HOG descriptor can separate the rain dictionary part from the whole dictionary, as already demonstrated in [1]. However, for different types of rain structures, it is not sure whether the handcrafted HOG descriptor can classify the rain dictionary from the whole dictionary. To check this point, rain images with various types of rain structures in size and shape are selected and then tested. Our experiments show that the sparse coding method [1] removes rain structures as well as object's structures. This indicates that the used prior knowledge is not always right, and thus the handcrafted HOG descriptor can fail to separate the rain dictionary from whole dictionary.

Similarly, in the discriminative sparse coding method [18], initial rain dictionary is created by using the motion kernel with a dominant gradient orientation of input rain image. In this method, it is assumed that rain structures are close to straight lines. However, there are various types of rain patterns in size and shape in the real world. As pointed out in [18], this discriminative sparse coding method is not applicable to rain images with magnified rain drops. In our experiment, similar results are obtained. Even though object's details can be preserved with the learned non-rain dictionary, rain structures still remain in the rain regions due to the initialized rain dictionary with improper prior knowledge. Therefore, the conventional sparse coding methods [1,18] needs to improve prior models more robust to various types of rain patterns, whereas in the proposed method, various types of real-life rain structures can be modeled by using the representation learning approach [15,14] with masked rain image database.

Second, in the proposed method, a new shrinkage-based approach is adopted to remedy edge artifacts and detail loss in non-rain regions and to remove rain structures in rain regions, which are the main issues of this paper, as already discussed with Fig. 1. The proposed shrinkage-based sparse coding determines how much the sparse codes of the rain and non-rain dictionaries are attenuated in rain and non-rain regions. The learned rain dictionary can reconstruct the rain patches, and thus the amounts of attenuations for the sparse codes of the rain and non-rain dictionaries can be determined based on representation errors between input rain patches and reconstructed rain patches. Also, correlation strength between the signal-atoms in the rain and non-rain dictionaries can be used to determine which sparse codes of the signal-atoms in the non-rain dictionary should be shrunk in the rain regions. Even though an iterative-based approach [18] can be used to modify the sparse codes of the rain and non-rain dictionaries, a challenging non-convex optimization problem should be solved, as pointed out by the authors. We found in the experiments that pixel saturation and clipping artifacts appear on the rain-removed images with the used greedy pursuit algorithm in [18]. Even though multi-block alternating optimization technique is used instead of the greedy pursuit algorithm, in this case, the discriminative sparse coding becomes too slow to converge for true solutions, as already mentioned in [18]. It is thus hard to find a desirable solution with fast convergence for the non-convex optimization problem. Moreover, as indicated in the previous paragraph, the iterative approach in [18] should initialize the rain dictionary. However, this initialization needs to be more accurate with improved prior modeling for various types of rain patterns.

Third, more comparisons between the proposed method and the conventional sparse coding methods [1,18] are provided in the experimental result section to show obvious advantages of the proposed method. Our experiments show that the proposed method is much stronger at the representation for textures and image structures (e.g., face, shirt' line) on the rain-removed images in comparison with the conventional methods [1,7]. Also, the proposed method is more effective in removing rain structures from similar image structures than the conventional methods [1,7].

This paper is the updated version of our conference paper [27]. Compared to the previous work [27], quantitative evaluation is newly added in this paper and more test images are compared. In addition, more discussion about the proposed method and the conventional methods [1,7,18] are provided.

## II. OUR RAIN REMOVAL MODEL

As mentioned in the Introduction, our main goal is to shrink the sparse codes of the rain and non-rain dictionaries. To achieve this, a shrinkage map normalized to [0-1] will be designed, and then multiplied to the sparse codes of the input rain patches. Our rain removal model is expressed, as follows:

$$\mathbf{R}_i \mathbf{x} \approx \mathbf{D} f(\mathbf{\alpha}_i) = \left[\mathbf{D}^n \mathbf{D}^r\right] f\left(\left[\mathbf{\alpha}_i^n ; \mathbf{\alpha}_i^r\right]\right) = \left[\mathbf{D}^n \mathbf{D}^r\right]\left(\mathbf{\alpha}_i^s = s_i \cdot \left[\mathbf{\alpha}_i^n ; \mathbf{\alpha}_i^r\right]\right) \quad (2)$$

where $f$ indicates the shrinkage function and the symbol of semicolon $(;)$ is used to create a new row in the vector or matrix. Our model uses a simple linear shrinkage function $f(\mathbf{\alpha}_i) = s_i \mathbf{\alpha}_i$. Where $s_i$ is a scalar value to shrink the sparse codes. The above equation shows that the input rain patch $\mathbf{R}_i \mathbf{x}$ will be replaced by the linear combination of the dictionary set $\mathbf{D}$ and the shrunk sparse codes $\mathbf{\alpha}_i^s$ to obtain the rain-removed patch.

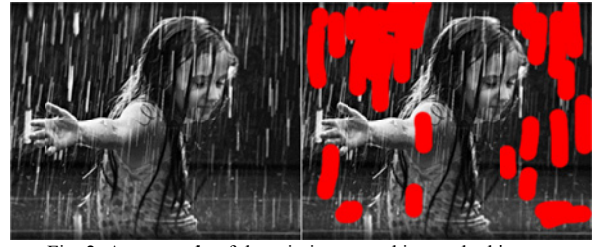
Fig. 2. An example of the rain image and its masked image.

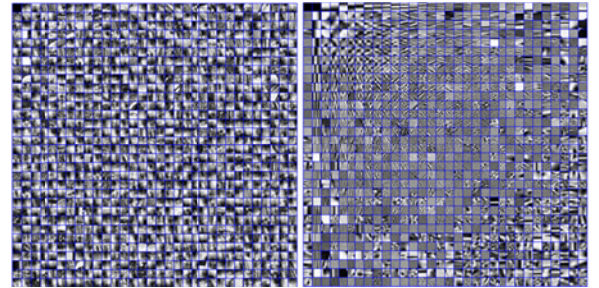
Fig. 3. Rain dictionary (left image) vs. non-rain dictionary (right image).

## III. PROPOSED RAIN REMOVAL METHOD

### A. Our Shrinkage Strategy

The purpose of using the shrinkage function in (2) is to remove the rain structures in the rain regions and to preserve the details of objects without any edge artifacts in the non-rain regions. To achieve this goal, for the non-rain regions, higher values should be assigned to $s_i$. This makes the sparse codes of the $\mathbf{\alpha}_i^r$ and $\mathbf{\alpha}_i^n$ change little in the non-rain regions, thereby avoiding the edge artifacts and preserving the details of objects. For the rain regions, lower values should be assigned to $s_i$, which can attenuate the magnitude of the sparse code $\mathbf{\alpha}_i^r$. As a result, the rain structures in the rain regions can be removed. However, there is a question about how to shrink the $\mathbf{\alpha}_i^n$ in the rain regions. Even in the rain regions, the highly correlated signal-atoms of the rain and non-rain dictionaries can be used jointly to represent the rain structures. In this case, the highly correlated signal-atoms should be removed in the rain regions. This can done by shrinking some parts of the sparse code $\mathbf{\alpha}_i^n$ with small value of $s_i$. In other words, some vector elements of the $\mathbf{\alpha}_i^n$, not the whole vector $\mathbf{\alpha}_i^n$, will be shrunk based on the correlation matrix between the two dictionaries $\mathbf{D}^n$ and $\mathbf{D}^r$ in the rain regions.

### B. Rain Image Database Construction

To learn the rain dictionary, a collection of rain images is needed. Moreover, as mentioned in the Introduction, the masked images are needed to indicate the rain regions in the rain images. In this paper, ninety rain images are downloaded from the websites, and then the masked images are generated manually. Fig. 2 shows the example of the rain image and the corresponding masked image. The red colors in the masked image indicate the rain regions. By using the masked images, rain patches are extracted via a random sampling. The total number of the extracted rain patches is 15000, which are used for the rain dictionary learning. It is thus expected that the learned rain dictionary can represent various types of rain structures, which makes it possible to detect rain regions. Note that the non-rain dictionary is trained by another image database that contains three hundred natural images without rain structures, which are also collected from websites.

One of the original papers that deals with the dictionary learning [15] shows that satisfactory patch sparsity and representation accuracy can be obtained with 11000 natural patches. Based on this guideline, 15000 rain patches are used in this paper for rain dictionary learning. As more rain images with different rain structures are added to our rain database, the representation accuracy of the learned rain dictionary can be strengthened.

### C. Offline Dictionary Learning

To learn the two types of rain and non-rain dictionaries ( $\mathbf{D}^n \in \Re^{m^2 \times K}$ and $\mathbf{D}^r \in \Re^{m^2 \times K}$ ), well-known K-SVD dictionary learning algorithm [15] was used. The dictionary size is $m^2 \times K = 256 \times 1024$. That is, the used patch size during the dictionary learning is $16 \times 16$ and the number of the signal-atoms (i.e., the column vectors in each dictionary) is 1024. For each rain and non-rain training database, dictionary learning was conducted separately.

Fig. 3 shows the example of the learned rain and non-rain dictionaries. In the images, each small square indicates the signal-atoms in the dictionary. For visualization, the signal-atoms are reshaped as blocks. It can be observed that the shape of the rain dictionary is not limited only to vertical and diagonal lines, and thus it can represent various types of rain patterns. Actually, the used dictionary learning is conducted based on patch unit, and therefore, even the input rain patches can include the structures of background and objects. In contrast, the non-rain dictionary has more variety of patterns to represent all natural patches.

### D. Shrinkage Map Design

To shrink the sparse code $\boldsymbol{\alpha}_i^r$ of the rain dictionary, a shrink map filled with the values of $s_i$ is first designed. This map is normalized to [0-1] and it has the same size as the input rain image. The shrinkage map is generated according to the following steps:

**Step 1:** Conduct the sparse coding using only the rain dictionary for all overlapping input rain patches extracted from input rain image. The orthogonal matching pursuit algorithm [12,13] is used for the sparse coding.

$$\min_{\boldsymbol{\alpha}_i} \|\mathbf{R}_i \mathbf{x} - \mathbf{D}^r \boldsymbol{\alpha}_i^r\| \text{ subject to } \|\boldsymbol{\alpha}_i^r\|_0 \leq L \quad (3)$$

where $\|\cdot\|_p$ indicates the $p$-norm and $L$ is a scalar value to control the sparsity. Note that the sparse code of $\boldsymbol{\alpha}_i^r$ estimated with the rain dictionary is not same as the one in (2), which is estimated using two types of the non-rain and rain dictionaries. During the sparse coding, $L$ is set to 3.

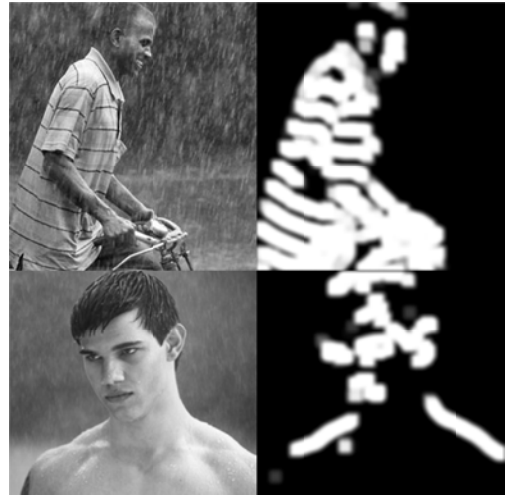
Fig. 4. Input rain images (left column) and their shrinkage maps (right column).

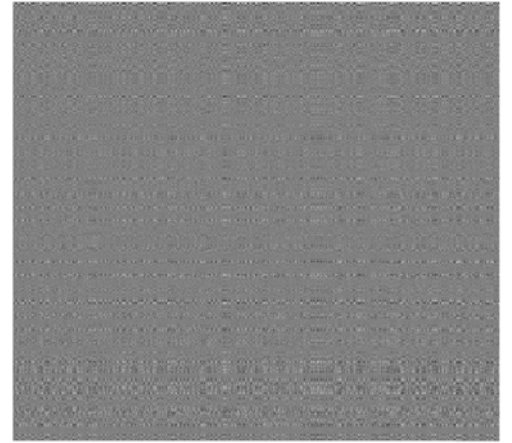
Fig. 5. Correlation matrix between the signal-atoms of the rain and non-rain dictionaries.

**Step 2:** Complete the whole image by averaging all the reconstructed patches ( $\mathbf{D}^r \boldsymbol{\alpha}_i^r$ ), and then generate the error map that includes the square of the difference between the whole image and the input rain image at each pixel position.

$$\mathbf{x}^* = \left( \sum_{i=1}^{N} \mathbf{R}_i^T \mathbf{R}_i \right)^{-1} \left( \sum_{i=1}^{N} \mathbf{R}_i^T \mathbf{D} \boldsymbol{\alpha}_i^r \right) \quad (4)$$

$$\mathbf{e}_j = (\mathbf{x}_j^* - \mathbf{x}_j)^2 \quad (5)$$

The patching averaging of the reconstructed overlapping patches is expressed by (4), which is derived from $\min_{\mathbf{x}} \sum_{i}^{N} \|\mathbf{R}_i \mathbf{x} - \mathbf{D}^r \boldsymbol{\alpha}_i^r\|_2^2$. In (4), $T$ is the transpose operator and $N$ is the total number of the overlapping patches. Please refer to [24,25] for more details of the patch averaging. As shown in (4), the rain dictionary is used to represent the input rain image, and thus it is expected that the reconstructed image $\mathbf{x}^*$ can represent the rain regions, however it cannot represent the non-rain regions. For this reason, the error map $\mathbf{e}$ in (5) will have large values for the non-rain regions, whereas it will have small values for the rain regions. In (5), $j$ indicates the pixel index in an image vector, which is not the same as the patch index $i$ in (4). In the Step 1 and 2, the learned non-rain dictionary can be used to increase the accuracy of the error map. However, this can increase computational cost. As shown in the experimental results, only use of the learned rain dictionary can provide satisfactory results. Thus, the non-rain dictionary was excluded in designing the error map.

**Step 3:** Apply the k-means clustering algorithm [28] to the error map, and then generate the shrinkage map based on the distance ratios between the clustered centers and pixel values in the error map.

$$\mathbf{s}_j = d_j^c / (d_j^{c_1} + d_j^{c_2}) \quad \text{and} \quad c = \min(c_1, c_2) \qquad (6)$$

where $\mathbf{s}$ is the shrinkage map; $c_1$ and $c_2$ are the cluster centers with scalar values. $d_j^{c_1}$ and $d_j^{c_2}$ are the distances between the $j$th pixel value in the error map and the cluster centers of the $c_1$ and $c_2$, respectively. In (6), $c = \min(c_1, c_2)$ indicates the cluster center for the rain regions. Therefore, the shrinkage map will have small values for the rain regions because the pixel values of the error map for the rain regions will be close to the cluster center ($c$). On the other hand, for the non-rain regions, the shrinkage map will have higher values.

**Step 4:** Find the pixels with horizontal lines in the input rain image, and then assign higher values (e.g., '1') to the same pixels in the shrinkage map. It can be assumed that rain structures rarely have horizontal edges. Finally, a gray dilation [15] is conducted on the shrinkage map to expand the non-rain regions. Visual saliency map [4] or DOF [17] would be considered to be added into the final shrinkage map. Also, rain regions detected by a trained classifier (e.g., support vector machine or deep learning) can be added to the final shrinkage map. More details about how to train classifiers are provided in the supplementary material. In this paper, Prewitt edge operator [29] was used to find the horizontal lines.

Fig. 4 shows the example of the generated shrinkage map, according to the four steps mentioned above. In this map, it can be observed that the non-rain regions have higher intensity values, whereas the rain regions have smaller intensity values. Therefore, this shrinkage map satisfies our shrinkage strategy, as mentioned in the subsection of III.A. In Fig. 4, the shrinkage map was scaled to [0-255] for visualization.

*E. Shrinkage of Rain Sparse Codes*

Given the shrinkage map, the sparse code $\boldsymbol{\alpha}_i^r$ of the rain dictionary will be shrunk with the proposed rain removal model, as shown in (2). In other words, the sparse code $\boldsymbol{\alpha}_i^r$ is multiplied with the corresponding shrinkage value $s_i$. However, the proposed rain removal method is conducted based on the patch unit. In other words, the index $i$ indicates the patch extracted from the input rain image at the $i$th pixel location. Actually, each sparse code $\boldsymbol{\alpha}_i^r$ corresponds to the extracted input rain patch, and thus the shrinkage value $s_i$ should be calculated from the extracted shrinkage patch at the same pixel location. For the shrinkage of the rain sparse codes, first, the sparse coding is conducted using the dictionary set, as follows:

$$\min_{\boldsymbol{\alpha}_i} \|\boldsymbol{\alpha}_i\|_0 \quad \text{subject to} \quad \|\mathbf{R}_i \mathbf{x} - \mathbf{D}\boldsymbol{\alpha}_i\| \le m^2 (\varepsilon)^2 \qquad (7)$$

Note that the dictionary set $\mathbf{D}$ and the sparse code $\boldsymbol{\alpha}_i$, not the $\mathbf{D}^r$ and $\boldsymbol{\alpha}_i^r$, are used in (7), according to the proposed rain removal model as defined in (2). To minimize (7), orthogonal matching pursuit was used [15]. In (7), $m^2 = 256$ is the dimension of the extracted patch $\mathbf{R}_i \mathbf{x}$ and $\varepsilon$ is the bounded representation error [15,25]. Discussion about how to set the bounded representation error is provided in supplementary material. Next, the shrinkage value $s_i$ is calculated, as follows:

$$s_i = f_{avg} (\mathbf{R}_i \mathbf{s}) \qquad (8)$$

where $\mathbf{s}$ is the shrinkage map and $f_{avg}$ is the average function. Thus, the value of $s_i$ is the mean of the extracted shrinkage patch $\mathbf{R}_i \mathbf{s}$. Given the $s_i$ and $\boldsymbol{\alpha}_i = [\boldsymbol{\alpha}_i^n; \boldsymbol{\alpha}_i^r]$, the sparse code $\boldsymbol{\alpha}_i^r$ corresponding to the rain dictionary is extracted from the $\boldsymbol{\alpha}_i$, and then shrunk as:

$$\boldsymbol{\alpha}_i^{r_s} = s_i \cdot \boldsymbol{\alpha}_i^r \qquad (9)$$

Given the shrinkage map, as shown in Fig. 4, the magnitude of the $\boldsymbol{\alpha}_i^r$ for the rain regions can be reduced via (9) because the shrinkage map has small values for the rain regions. Also, the value of the $\boldsymbol{\alpha}_i^r$ can be preserved for the non-rain regions because the shrinkage map has higher values in those regions. Therefore, it is expected that the rain structures can be removed and the edge artifacts can be avoided. Moreover, the image structures of objects (e.g., face or lines on the shirt in Fig. 4) can be preserved.

*F. Shrinkage of Non-Rain Sparse Codes*

Now, we move on to the shrinkage of the non-rain sparse coding $\boldsymbol{\alpha}_i^n$. In rain regions, the sparse codes of rain and non-rain dictionaries can be used to reconstruct rain structures. Therefore, the signal-atoms of the non-rain dictionary that are highly correlated to the signal-atoms of the rain dictionary should be removed in the rain regions. However, if the sparse codes of the non-rain dictionary are forced to be zero vectors in the rain regions, over-smoothing effect can occur in the rain regions. This means that fine textures (e.g., tree leaves) in the rain regions, will be removed along with rain structures. Thus, in this paper, the signal-atoms of the non-rain dictionary that are highly correlated to the signal-atoms of the rain dictionary are removed. To achieve this, the correlation matrix is needed to know how much the signal-atoms of the non-rain and rain dictionaries are correlated. As mentioned in the shrinkage strategy in the subsection of III.A, some elements of the $\boldsymbol{\alpha}_i^n$, not the whole vector $\boldsymbol{\alpha}_i^n$, will be shrunk based on the correlation matrix between the two dictionaries $\mathbf{D}^n$ and $\mathbf{D}^r$. The correlation matrix [30] is calculated as follows:

$$\mathbf{C}(k,l) = \frac{\mathbf{D}^n(:,k)^T \mathbf{D}^r(:,l)}{\sqrt{\mathbf{D}^n(:,k)^T \mathbf{D}^n(:,k)} \sqrt{\mathbf{D}^r(:,l)^T \mathbf{D}^r(:,l)}} \qquad (10)$$

where $\mathbf{C}$ is the correlation matrix and $(k,l)$ is the index to indicate the matrix elements. $\mathbf{D}^n(:,k)$ and $\mathbf{D}^r(:,l)$ is the $k$th and $l$th column vectors of the $\mathbf{D}^n$ and $\mathbf{D}^r$, respectively. The correlation matrix can measure how much the two signal-atoms, i.e., $\mathbf{D}^n(:,k)$ and $\mathbf{D}^r(:,l)$ are similar.

Fig. 5 shows the correlation matrix calculated from the signal-atoms of the learned two dictionaries, which is shown in Fig. 3. The size of the matrix $\mathbf{C}$ is $1024 \times 1024$ because the total number of the signal-atoms in each dictionary is 1024. In the correlation matrix, the total number of the signal-atoms in the dictionary $\mathbf{D}^n$ that have the correlation values higher than a threshold $TH_c = 0.8$ is 42. In other words, 42 out of 1024 signal-atoms in the non-rain dictionary are similar to the ones in the rain dictionary $\mathbf{D}^r$. This means that some signal-atoms of the non-rain dictionary can be used to represent the rain regions with the corresponding highly correlated signal-atoms in the rain dictionary. Therefore, in the rain regions, the highly correlated signal-atoms of the rain and non-rain dictionaries should be removed at the same time. This can be done by shrinking the sparse codes of the $\boldsymbol{\alpha}_i^n$ and $\boldsymbol{\alpha}_i^r$. In the rain region, as already mentioned in the previous section, the sparse code of the $\boldsymbol{\alpha}_i^r$ can be shrunk, according to (7)-(9). The remaining sparse code of the $\boldsymbol{\alpha}_i^n$ will be shrunk, as follows:

$$\boldsymbol{\alpha}_i^{n_s}(k) = s_i \boldsymbol{\alpha}_i^n(k) \text{ if } s_i \leq TH_s, \; \boldsymbol{\alpha}_i^r(l) \neq 0, \text{ and } C(k,l) \geq TH_c \quad (11)$$

where $\boldsymbol{\alpha}_i^n(k)$ indicates the $k$th element of the vector $\boldsymbol{\alpha}_i^n$. Therefore, above equation shows that some elements of the $\boldsymbol{\alpha}_i^n$ are shrunk with the value of $s_i$. However, there are some constraints: $s_i \leq TH_s$, $\boldsymbol{\alpha}_i^r(l) \neq 0$, and $C(k,l) \geq TH_c$. In (11), the first constraint $s_i \leq TH_s$ indicates that the input rain patch should belong to the rain regions. The second and third constraints indicate that the signal-atom $\mathbf{D}^n(:,k)$ should be highly correlated to the signal-atom $\mathbf{D}^r(:,l)$. Also, this signal-atom $\mathbf{D}^r(:,l)$ should have non-zero sparse coefficient, i.e., $\boldsymbol{\alpha}_i^r(l) \neq 0$. Consequently, by doing (9) and (11), in the rain regions, the influences of the highly correlated signal-atoms of the rain and non-rain dictionaries can be removed at the same time. This can lead to improvement in rain removal, especially for the rain regions. In (11), the parameters are heuristically set by $TH_c = 0.8$ and $TH_s = 0.25$.

---

**Algorithm I: Proposed rain removal**

**Input**: Rain image $\mathbf{x}$ and learned dictionary set $\mathbf{D} = [\mathbf{D}^n \mathbf{D}^r]$

**Output**: Rain-removed image $\mathbf{y}$

**Initialization**:
- Generate the shrinkage map $\mathbf{s}$, according to **Step** 1-4
- Find the correlation matrix $\mathbf{C}$ using (10)
- Initialize the parameter $\varepsilon$, $L$, $TH_s$, and $TH_c$ used in (7), (11), and (3)
- Initialize the vector $\mathbf{p}$ and the matrix $\mathbf{Q}$ by all zeros

**Proposed Rain Removal**:

**for** $i = 1, 2, ..., N$
- Conduct the orthogonal matching pursuit via (7) for each input patch $\mathbf{R}_i \mathbf{x}$
- Calculate the shrinkage value $s_i$ using (8)
- Obtain the shrunk sparse code $\boldsymbol{\alpha}_i^{r_s}$, according to (9)
- Obtain the shrunk sparse code $\boldsymbol{\alpha}_i^{n_s}$, according to (11)
- Make the rain-removed patch via $\mathbf{D}\boldsymbol{\alpha}_i^s = [\mathbf{D}^n \mathbf{D}^r][\boldsymbol{\alpha}_i^{n_s}; \boldsymbol{\alpha}_i^{r_s}]$
- Update the $\mathbf{p} = \mathbf{p} + \mathbf{R}_i^T \mathbf{D}\boldsymbol{\alpha}_i^s$ and $\mathbf{Q} = \mathbf{Q} + \mathbf{R}_i^T \mathbf{R}_i$

**end**
- Conduct the patch averaging to obtain the rain-removed image $\mathbf{y} = \mathbf{p} / diag(\mathbf{Q})$ where $diag$ is the function to extract the diagonal elements from a matrix and then make a column vector

**Return** $\mathbf{y}$

---

### G. Proposed Rain Removal Algorithm

In Algorithm I, the proposed rain removal algorithm is summarized. After initializing the shrinkage map $\mathbf{s}$, correlation matrix $\mathbf{C}$, and parameters ($\varepsilon$, $L = 3$, $TH_s = 0.25$, $TH_c = 0.8$), orthogonal matching pursuit [15] is conducted for every overlapping patch $\mathbf{R}_i \mathbf{x}$ extracted from the input rain image $\mathbf{x}$. Then, the shrinkage value of $s_i$ is calculated from the shrinkage map using (8), and the sparse codes $\boldsymbol{\alpha}_i^r$ and $\boldsymbol{\alpha}_i^n$ are shrunk using (9) and (11), respectively. Next, the rain-removed patch is generated by the linear combination of the dictionary set $\mathbf{D}$ and the shrunk sparse code $\boldsymbol{\alpha}_i^s$. Then, the vector $\mathbf{p}$ and the matrix $\mathbf{Q}$ are updated to save the reconstructed patches and the number of the overlapping patches. After finishing the sparse coding for all patches, patch averaging is conducted to obtain the rain-removed image $\mathbf{y}$ via $\mathbf{p} / diag(\mathbf{Q})$, where $diag$ is the function to extract the diagonal elements from a matrix and then make a column vector.

### IV. EXPERIMENTAL RESULTS

Three experiments will be conducted in this section. First, it will be shown how the proposed shrinkage approach for each $\boldsymbol{\alpha}_i^r$ and $\boldsymbol{\alpha}_i^n$ can affect visual effects, and then visual comparison between the proposed method and the conventional methods [1,7,18] will be given. Finally, image quality evaluation will be conducted. The used rain image database and Matlab source code will be uploaded at the website: https://sites.google.com/site/changhwan76son/

## A. Visual Effects According to Proposed Shrinkage Model

Fig. 6(a) shows the visual effect according to the proposed shrinkage approach for the sparse code $\alpha_i^r$. In Fig. 6(a), the left image shows the rain-removed image with the conventional shrinkage approach ($\alpha_i^r = 0$) [1], whereas the right image shows the rain-removed image with the proposed shrinkage model ($s_i\alpha_i^r$). In the left image, as already discussed in the Introduction, the edge artifacts are generated around the man's shoulder. Also, contrast and detail loss occurred in the face regions. However, the use of the proposed shrinkage approach modeled with $s_i\alpha_i^r$ can reduce the edge artifacts and improve the contrast and details of the face, as shown in right image. As shown in Fig. 4, the shrink map has higher values around the man's shoulder. Therefore, in those regions, the sparse codes of the $\alpha_i^r$ can be preserved, i.e., $s_i\alpha_i^r \approx \alpha_i^r$. This means that the spare code $\alpha_i^r$ of the rain dictionary can be used to represent the structures of the man's shoulder with another sparse code $\alpha_i^n$ of the non-rain dictionary. This can avoid the edge artifacts and preserve the image structures at the same time.

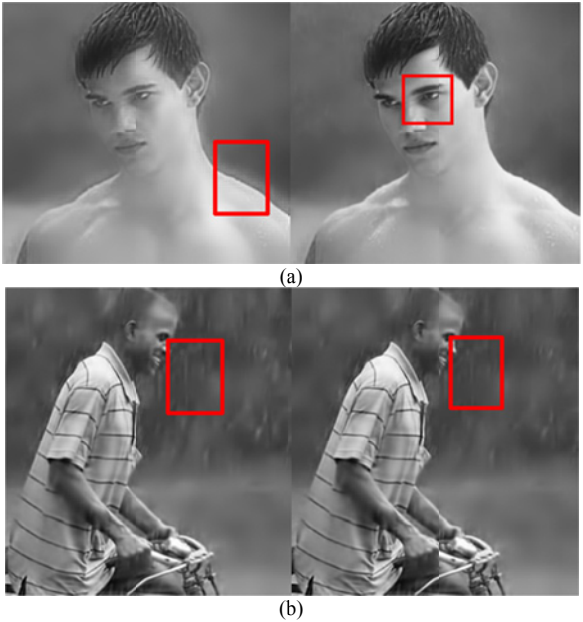

(a)

(b)

Fig. 6. Visual effects, according to the proposed shrinkage: (a) rain-removed image with the $\alpha_i^r = 0$ (left image) and rain-removed image with the proposed shrinkage model $s_i\alpha_i^r$ (right image) and (b) rain-removed images without any shrinkage of the $\alpha_i^n$ (left image) and rain-removed image after applying the proposed shrinkage to the $\alpha_i^n$ (right image).

Fig. 6(b) shows the visual effect according to the use of the proposed shrinkage approach for the sparse code $\alpha_i^n$. As it is shown in the red boxes, it can be observed that the rain structures can be suppressed more with the proposed shrinkage approach, as defined in (11). This reveals that a part of the signal-atoms in the non-rain dictionary can be employed to represent the rain structures (or rain regions) with the corresponding highly correlated signal-atoms in the rain dictionary. Therefore, in the rain regions, the highly correlated signal-atoms of the rain and non-rain dictionaries should be removed at the same time. The proposed shrinkage approach for the $\alpha_i^n$ based on the correlation matrix can remove the signal-atoms of the non-rain dictionary that are highly correlated to the rain dictionary, thereby improving the rain removal, especially for the rain regions.

## B. Visual Quality Comparison

Figs. 7-12 shows the rain-removed images using the conventional methods [1,7] and the proposed method. Here, the test images, as shown in Figs. 7(a), 8(a), and 9(a), were used as training images for rain dictionary learning. However, other test images, as shown in Figs. 10-12, are not used as the training images. As shown in Figs. 7(b) and 11(b), the texture removal method using RTV [7] can be a good solution if the rain structures are fine. However, if the rain structures are thick, the RTV cannot discriminate between the rain and the image structures, and thus the texture removal method can remove the image structures and rain structures simultaneously, as shown in Figs. 9(b) and 10(b). The conventional sparse coding method [1] can remove the rain structures well. Especially, the HOG descriptor is strong at the representation for thick and long rain steaks, and thus the sparse coding method can remove the long rain steaks, as shown in the red box of Fig. 8(c). For the long rain steaks, the rain removal performance of the sparse coding method is better than the proposed method, as shown in Figs. 8(c) and 8(d). However, the HOG descriptor can suffer from separating the rain dictionary from the whole dictionary for different types of rain patterns. As a result, the details of the tree leaves and face are almost removed, as shown in Figs. 7(c), 8(c) and 11(c). In contrast, the proposed rain removal method collects the rain structures using the masked images, as shown in Fig. 2, and then learns the rain structures via offline dictionary learning. As mentioned in Introduction, recent trend is to adopt the learning representation. It can be a better choice to use the learned rain dictionary rather than the handcrafted HOG features to represent various types of rain patterns with different size and shape. Taking advantage of the learned rain dictionary, which can represent the rain structures accurately, but it cannot represent the image structures of objects, an error map can be generated. Based on the shrinkage map induced from the error map, the sparse codes of the rain and non-rain dictionaries can be used to represent the image structures in non-rain regions. As a result, the image structures (e.g., face and tree leaves) in the non-rain regions can be described more accurately with the proposed method than the conventional methods [1,7]. Especially, the proposed method can distinguish the rain structures from the raindrops. This can be checked in the red boxes of the Fig. 11d and the third row of Fig. 12, respectively, where the rain structures are removed, however the raindrops falling on the ground or face can be preserved. In addition, rain structures can be more suppressed with the proposed method even though there are still a few rain structures on the rain regions. The use of the correlation matrix enables the highly correlated signal-atoms of the rain and non-rain dictionaries to be removed in the rain regions, and thus

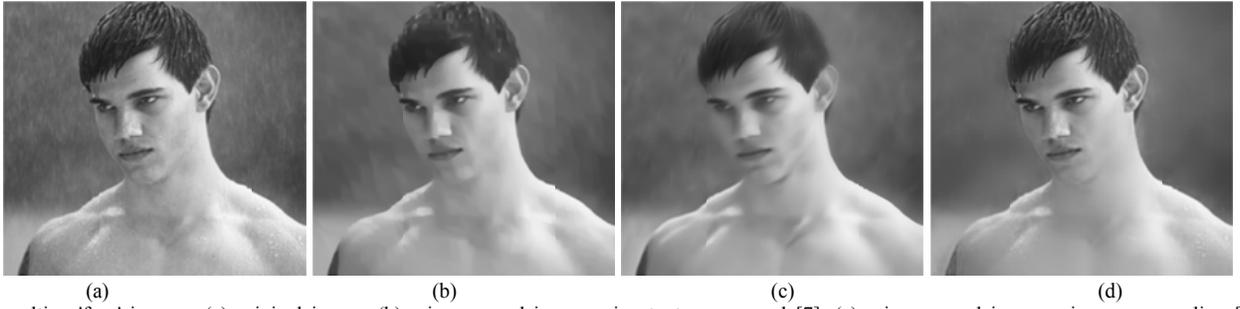

Fig. 7. Resulting 'face' images; (a) original image, (b) rain-removed image using texture removal [7], (c) rain-removed image using sparse coding [1], (d) rain-removed image using the proposed method.

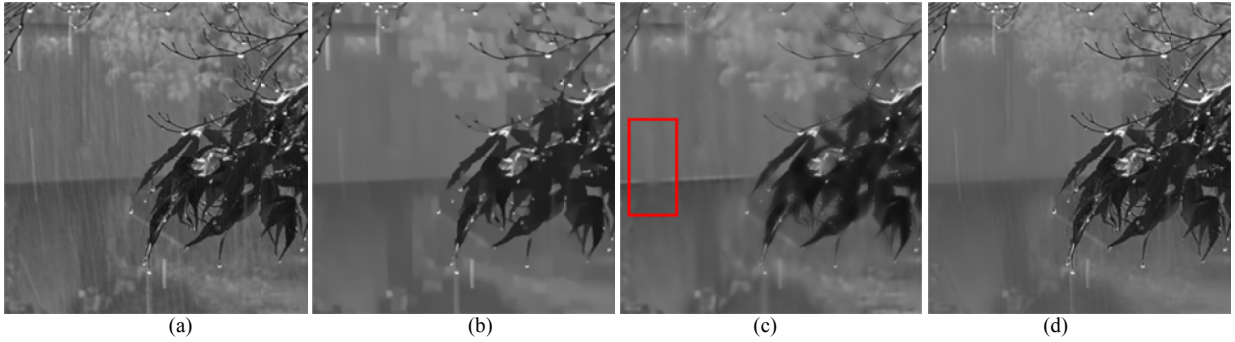

Fig. 8. Resulting 'tree' images; (a) original image, (b) rain-removed image using texture removal [7], (c) rain-removed image using sparse coding [1], (d) rain-removed image using the proposed method.

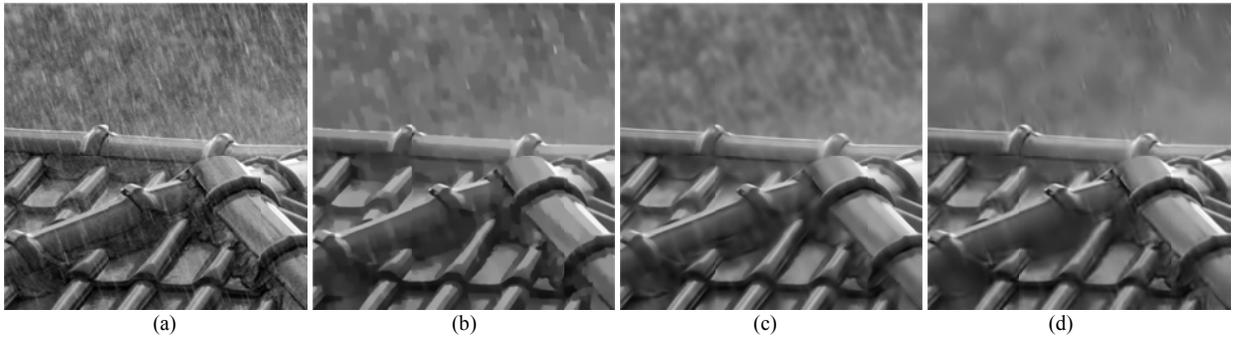

Fig. 9. Resulting 'roof tile' images; (a) original image, (b) rain-removed image using texture removal [7], (c) rain-removed image using sparse coding [1], (d) rain-removed image using the proposed method.

rain structures can be more suppressed. In addition, from the resulting images of Figs. 10-11, it can be said that the learned rain dictionary can detect other types of rain patterns that are not included in the training rain images.

In this paper, supplementary material is additionally provided to check the performance of the rain removal via discriminative sparse coding [18]. As shown in the resulting images, object details can be preserved, thanks to the use of the learned non-rain dictionary, similarly to the proposed method. However, rain structures cannot be removed. As mentioned in the Introduction, the discriminative sparse coding method [18] needs to initialize the rain dictionary with a prior knowledge about rain structures. In this method, it is assumed that rain structures are close to straight lines. Based on this prior knowledge, the rain dictionary is initialized by using the motion kernel with a dominant gradient orientation of the input rain image. However, there are various types of rain patterns in size and shape in the real world. Therefore, this initialization can fail to remove various types of rain structures.

Moreover, the discriminative sparse coding method [18]

needs to solve a challenging non-convex optimization, as pointed out by the authors. Our experiment found that pixel saturation and clipping artifacts appear on the rain-removed images (see the supplementary material). Thus, the used greedy pursuit algorithm needs to be more stable and robust irrespective of input rain images. Even though multi-block alternating optimization technique is used instead of the greedy pursuit algorithm, the discriminative sparse coding method becomes too slow to converge for true solutions, as already indicated in [18]. It is hard to find a solution with fast convergence for non-convex optimization problem. For this reason, visual quality comparison and quantitative evaluation are excluded in this paper.

### C. Limitations and Future Work of Proposed Method

The proposed method has some drawbacks. First, in this paper, to avoid detail loss and edge artifacts, the shrinkage map was adopted. According to the shrinkage strategy, the shrinkage map should have large values around object's boundary. In other words, object's boundary is classified as non-rain regions. However, the proposed method is conducted based on the patch unit, and thus the rain structures around the object's boundaries can be preserved. This can be detected in the boundaries of the man and umbrella of Fig. 10d and Fig. 11d, respectively. However, it is not easy to solve this problem because accurate boundary detection is required from input rain images, and then the rain structures around object's boundary should be removed. Second, the representation accuracy of the learned rain dictionary depends on the used training rain patches. If input rain images contain rain structures that are not included in the training images, the proposed method may fail to remove the rain structures.

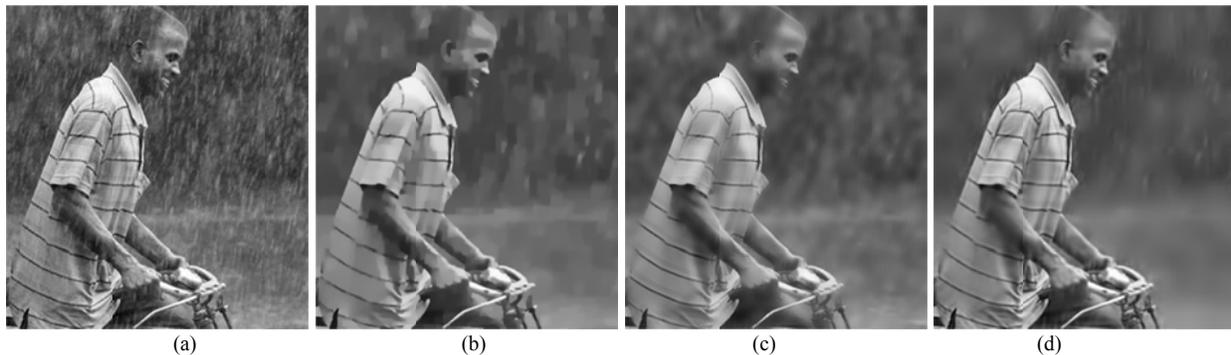

Fig. 10. Resulting 'man' images; (a) original image, (b) rain-removed image using texture removal [7], (c) rain-removed image using sparse coding [1], (d) rain-removed image using the proposed method.

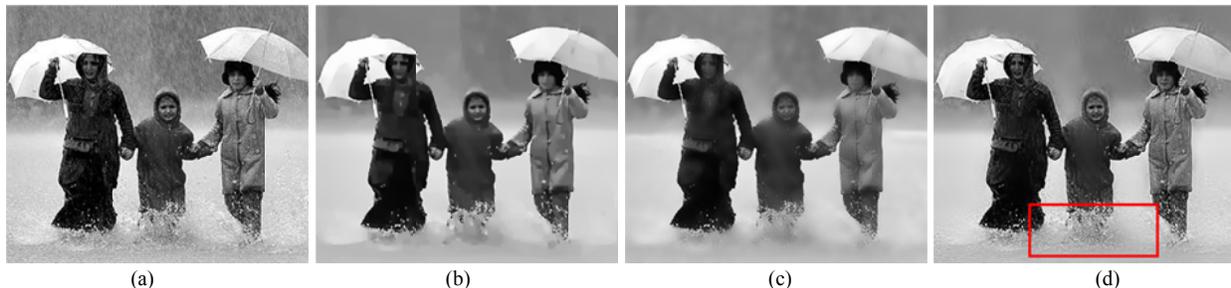

Fig. 11. Resulting 'people' images; (a) original image, (b) rain-removed image using texture removal [7], (c) rain-removed image using sparse coding [1], (d) rain-removed image using the proposed method.

There are still rooms for improvement in the proposed method. As shown in the red box of the last row of Fig. 12, it is a challenging task to remove the rain structures falling on the face. One solution is to learn color dictionaries from color rain images [31]. As shown in the red box, the rain structures falling on face or human body tend to have white colors, and thus color dictionaries can help to remove those rain structures. Moreover, rain structures can vary in size. In this paper, patch size is fixed, and thus the proposed method can fail to classify similar rain and image structures. To solve this problem, multi-resolution approach [32] can be considered to discriminate between image and rain structures at different scales. However, these issues will be handled in the future work.

### D. Quality Evaluation

To evaluate the performance of the proposed and conventional methods, blind image quality evaluation (BIQE) [33] and reference-based quality evaluations are considered for natural and synthetic rain images, respectively. First, to evaluate natural rain images, as shown in Figs. 7-12, opinion-*unaware* method [33] that does not require any human subjective scores for training is used. This method models the natural statistics of the local structures, contrast, multiscale decomposition, and then it measures the deviation of the distorted images from the reference statistics. In the rain-removed images, remaining rain structures can be considered as noise. Also, image structures can be removed after applying the rain removal. Therefore, the BIQE method can be used to measure how well the image structures can be preserved and how well the noisy rain structures are removed based on the natural image statistics. Table I shows the BIQE scores for the three methods introduced in the previous section. In Table I, BIQE scores become smaller when the natural statistic of the rain-removed image approaches to the reference natural statistic, which is determined using training images. As shown in Table I, the proposed method has the lowest average BIQE score. This means that the natural statistics of the rain-removed images using the proposed method are more close to the reference natural statistic than the conventional methods [1,7]. Thus, it can be deducted that the visual quality of the rain-removed images with the proposed method is better than the conventional methods. Also, this BIQE result confirms that the proposed method is effective at removing rain structures with small and moderate sizes and also in preserving object's details.

Second, to evaluate synthetic rain images, SSIM (Structure SIMilarity) [34] and PSNR (peak signal to noise ratio) are used. In this paper, to create synthetic rain images, rain patches are extracted from natural rain images, as shown in Figs. 7 and 12, and then added to original clean images. If it is necessary, the rain patches are rotated and then added to original clean images. Fig. 13 shows the created synthetic images. As shown in the synthetic 'brick' and 'face sketch' images, rain patches are chosen to have similar edge directions in the image structures of the horizontal and diagonal lines. For these synthetic images, we can check how effectively the proposed shrinkage-based sparse coding can remove rain structures and preserve image structures, compared to the conventional methods [1,7]. For the 'straw' image, we can check whether the proposed method using the learned rain dictionary can discriminate the rain structures from similar texture patterns but with different edge directions. The synthetic images, as shown in the second column, are also rotated 90 degrees, and then tested to check the performance of the rain removal methods for different rain directions.

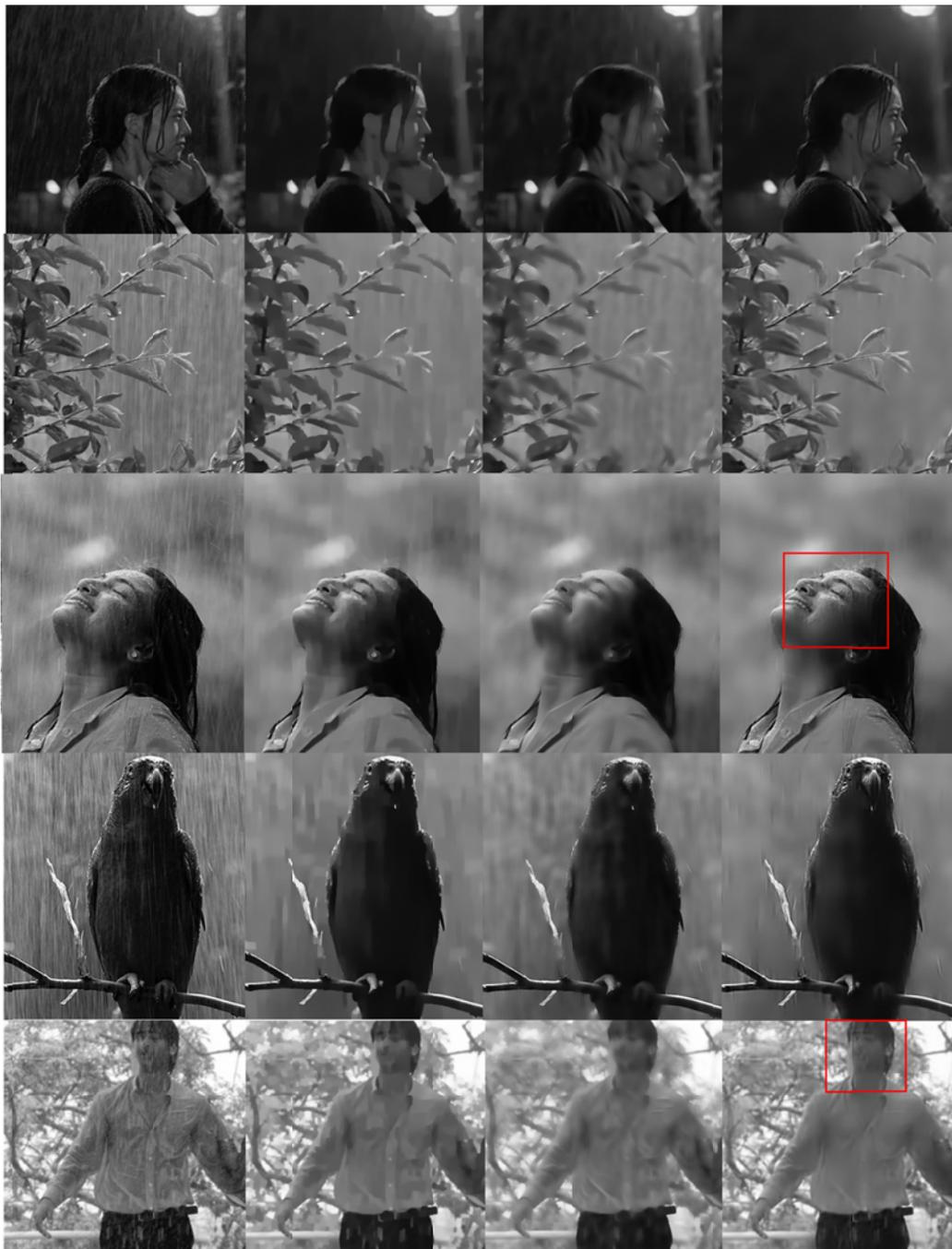

Fig. 12. Resulting images; original images (first column), rain-removed images using texture removal [7] (second column), rain-removed images using sparse coding [1] (third column), (d) rain-removed images using the proposed method (last column).

As shown in the shrinkage maps (last column), the proposed method can classify rain structures from image structures with similar edge directions. In the shrinkage map of the 'brick' image, brick textures are classified as non-rain regions, which are marked with white colors, whereas rain structures are classified as rain regions, which are marked with black colors. Similar results can be found in other shrinkage maps of the 'straw' and 'face sketch' images. As a result, in the 'brick' resulting image, the proposed method can preserve fine surface textures and horizontal lines while removing the rain structures. Also, diagonal lines on the 'face sketch' image are preserved, whereas the rain structures with similar edge directions can be removed. For the 'straw' image, textures can be distinguished from similar rain structures but with different edge directions. In contrast, the conventional sparse coding [1] and texture removal methods [7] remove the rain structures as well as image structures, as shown in blue boxes. For example, brick's surface textures, straw's textures, and diagonal lines are removed. In the case of the sparse coding method [1], face's details and contrasts are decreased. Also, the rain structures are not removed completely for the 'brick' and 'straw' images. This indicates that the HOG descriptor used in [1] cannot discriminate between image structures and rain structures with similar edge directions. Thus, the rain dictionary part cannot be accurately separated from the whole dictionary. Also, the RTV measure used in [7] cannot distinguish rain structures from fine textures. Thus, fine textures are removed on the rain-removed images.

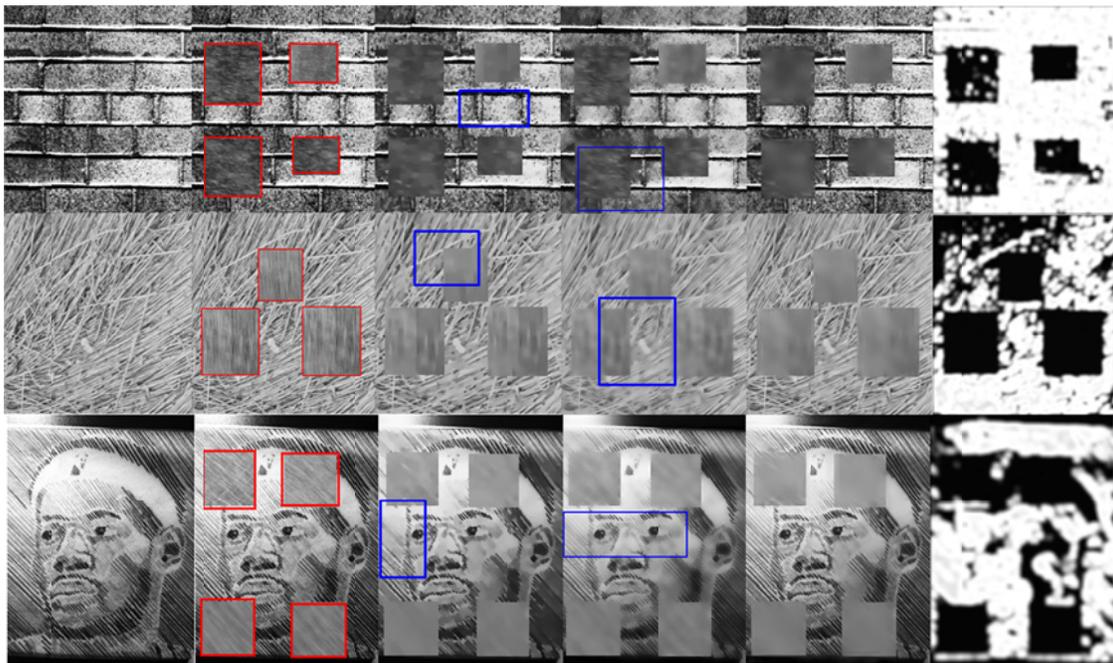

Fig. 13. Experimental results for synthetic images: original clean images of the 'brick', 'straw', and 'face sketch (first column), the created synthetic images (second column) where real-life rain patches are inserted into the red boxes, rain-removed images with texture removal [7] (third column), rain-removed images with sparse coding [1] (fourth column), and rain-removed images with proposed method (fifth column), and the estimated shrinkage maps with proposed method (last column).

TABLE I. BLIND IMAGE QUALITY EVALUATION

|  | Fig. 7 | Fig. 8 | Fig. 9 | Fig. 10 | Fig. 11 | Fig. 12 (1$^{st}$ row) | Fig. 12 (2$^{nd}$ row) | Fig. 12 (3$^{th}$ row) | Fig. 12 (4$^{rd}$ row) | Fig. 12 (5$^{th}$ row) | AVG. |
| --- | --- | --- | --- | --- | --- | --- | --- | --- | --- | --- | --- |
| Texture removal [7] | 55.404 | 44.860 | **52.539** | 45.159 | 40.002 | 51.882 | 49.643 | 44.645 | 49.343 | **48.792** | 48.226 |
| Sparse coding [1] | 65.582 | 44.516 | 59.812 | **39.815** | 34.835 | 63.987 | 51.499 | 53.055 | 41.4876 | 59.328 | 51.391 |
| Proposed method | **44.472** | **30.691** | 55.703 | 40.694 | **29.282** | **46.410** | **39.192** | **37.310** | **37.894** | 56.862 | **41.851** |

TABLE II. PSNR EVALUATION

|  | Brick | Brick (90) | Face sketch | Face sketch (90) | Straw | Straw (90) | AVG. |
| --- | --- | --- | --- | --- | --- | --- | --- |
| Texture removal [7] | 22.8612 | 22.6284 | 23.2503 | 22.6543 | 24.3136 | 24.8790 | 23.431 |
| Sparse coding [1] | 22.1515 | 22.2031 | 23.2932 | 23.3050 | 23.4263 | 22.1586 | 22.756 |
| Proposed method | **23.9983** | **25.2518** | **27.0618** | **26.4739** | **25.2232** | **25.7074** | **25.619** |

TABLE III. SSIM EVALUATION

|  | Brick | Brick (90) | Face sketch | Face sketch (90) | Straw | Straw (90) | AVG. |
| --- | --- | --- | --- | --- | --- | --- | --- |
| Texture removal [7] | 0.8157 | 0.8058 | 0.7111 | 0.6787 | 0.7494 | 0.7761 | 0.7561 |
| Sparse coding [1] | 0.8096 | 0.8115 | 0.7369 | 0.7364 | 0.7536 | 0.6650 | 0.7521 |
| Proposed method | **0.8911** | **0.9202** | **0.8831** | **0.8885** | **0.8285** | **0.8488** | **0.8767** |

Tables II and III show the PSNR and SSIM scores of the conventional and proposed methods for the synthetic images. In Tables II and III, round brackets indicate the rotation. As shown in these tables, the averaged PSNR and SSIM scores of the proposed method are higher than those of the conventional methods. This indicates that the proposed method is much stronger in representing object's details and textures on the rain-removed images than the conventional methods [1,7]. Also, this result shows that the proposed method is more effective at removing rain structures from similar image structures.

## V. CONCLUSIONS

A new rain removal model based on the shrinkage of the sparse codes is introduced in this paper. Direct use of the learned rain and non-rain dictionaries can generate unwanted edge artifacts and detail loss. This observation brought us to develop a new shrinkage-based sparse coding for rain removal. To realize this, in this paper, shrinkage map and correlation matrix were generated based on the learned rain and non-rain dictionaries. The shrinkage map can make the sparse codes of the rain and non-rain dictionaries change little in the non-rain regions, thereby avoiding edge artifacts and detail loss. In the rain regions, the correlation matrix can find the signal-atoms of the non-rain dictionary that are highly correlated to the ones in the rain dictionary so that the sparse codes corresponding to the non-rain dictionary can be shrunk in the rain regions. This leads to improvement in the rain removal, especially for the rain regions. Experimental results showed that the proposed rain removal model is good at preserving image structures and removing rain structures. Moreover, it is expected that the proposed rain removal model can be directly applied to snow removal if a snow image database is provided.

# Supplementary Material

## 1. Experimental results of the discriminative sparse coding method [18]

The discriminative sparse coding method [18] has been tested. The open source code at the author's website (Hui Ji) has been downloaded, and then tested with adjusting some parameters. The resulting images are given below,

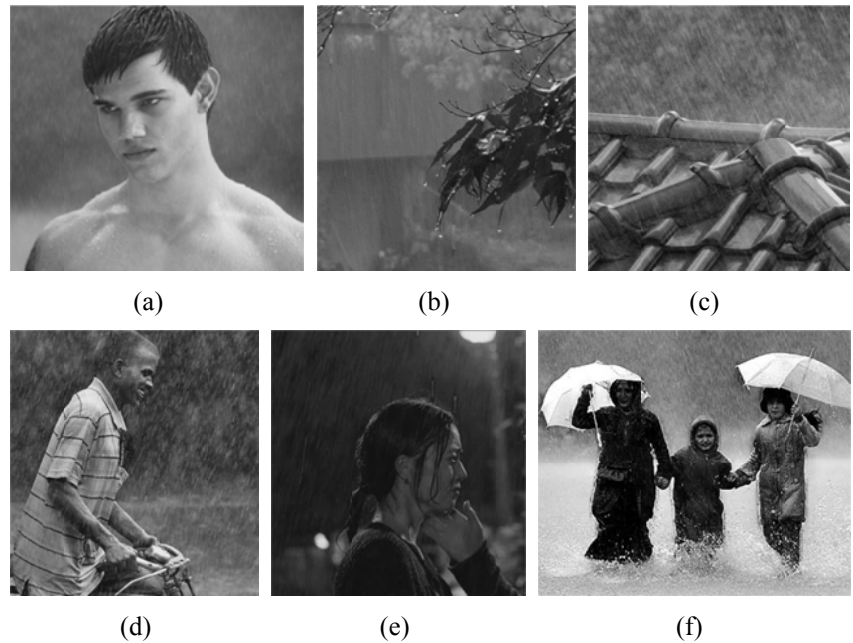

|  (a)  |  (b)  |  (c)  |
|  (d)  |  (e)  |  (f)  |

Fig. 14. Resulting images with the discriminative sparse coding method [18]

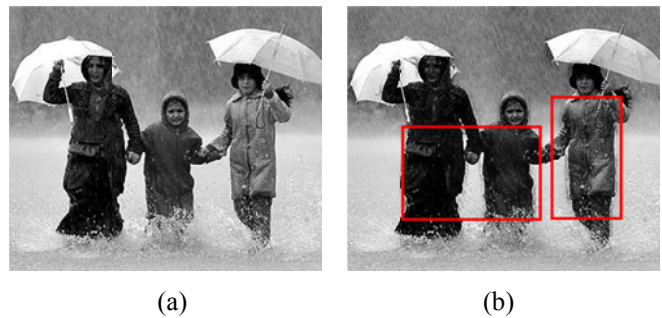

(a)                    (b)

Fig. 15. (a) Input rain image and (b) Resulting image with the discriminative sparse coding method [18]

As shown in Fig. 14, object details can be preserved, thanks to the use of the learned non-rain dictionary, similarly to the proposed method. However, rain structures cannot be removed. As mentioned in the previous page, the discriminative sparse coding method [18] needs to initialize the rain dictionary with a prior knowledge about rain structures. In this method, it is assumed that rain structures are close to straight lines. Based on this prior knowledge, the rain dictionary is initialized by using the motion kernel with a dominant gradient orientation of input rain image. However, there are various types of rain patterns in size and shape in the real world. Therefore, this initialization can fail to remove various types of rain structures, as shown in Fig. 14.

Moreover, the discriminative sparse coding method [18] needs to solve a challenging non-convex optimization, as pointed out by the authors. Our experiment found that pixel saturation and clipping artifacts appear on the rain-removed images. For example, in the red box of Fig. 15(b), pixel brightness becomes so dark, compared to the input rain image of Fig. 15(a). Thus, the used greedy pursuit algorithm needs to be more stable and robust irrespective of input rain images. Even though multi-block alternating optimization technique is used instead of the greedy pursuit algorithm, in this case, the discriminative sparse coding method becomes too slow to converge for true solutions, as already indicated in [18]. It is thus hard to find a solution with fast convergence for non-convex optimization problem.

In conclusion, as shown in resulting images of new experiments, the discriminative sparse coding method can fail to remove various types of rain patterns, due to the used initialization and greedy pursuit algorithm. On the other hands, our paper shows a new shrinkage-based sparse coding for rain removal, which is significantly different from the discriminative sparse coding method. It is shown that our shrinkage-based sparse coding approach is superior to the discriminative sparse coding method, as shown in the resulting images.

## 2. More comparisons between the proposed method and conventional methods [1,7]

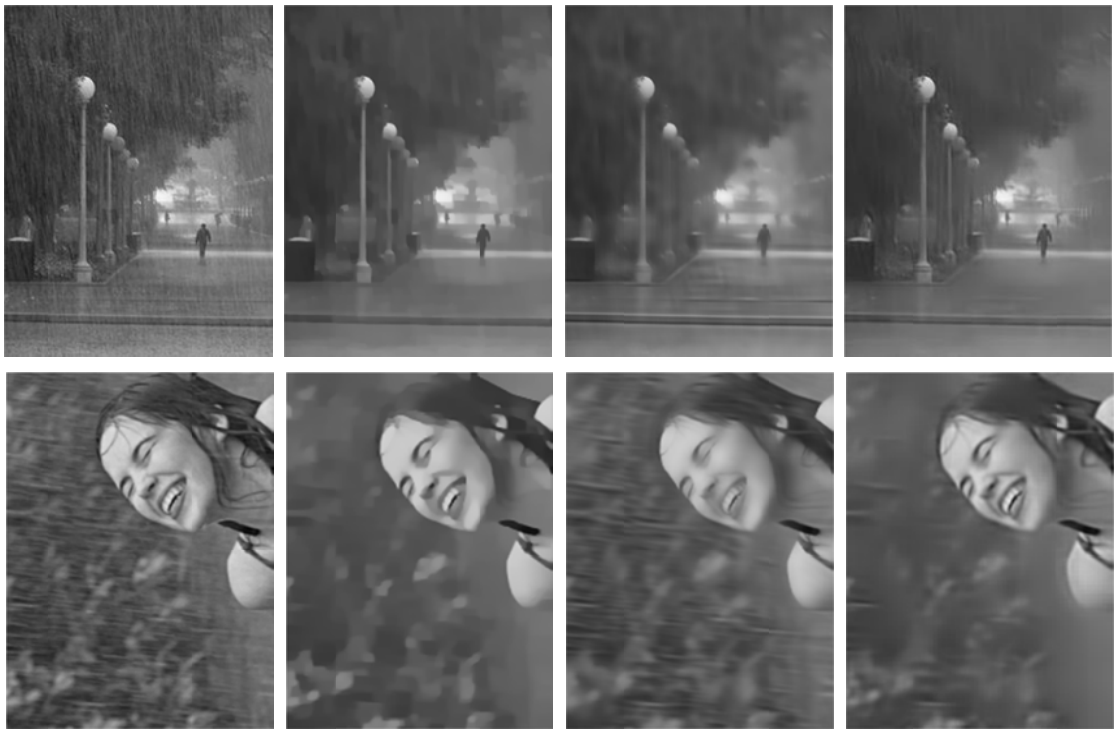

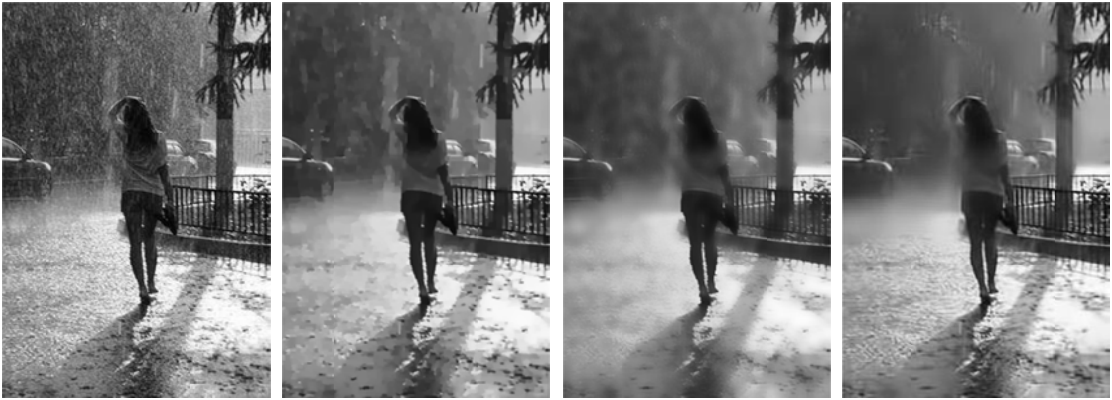

Fig. 16. Resulting images; original rain images (first column), rain-removed images using texture removal [7] (second column), rain-removed images using sparse coding [1] (third column), (d) rain-removed images using the proposed method (last column).

## 3. Parameter setting of the bounded representation error

The bounded representation error $\varepsilon$, as shown in (7), can be set manually or adaptively, according to input rain images. This bounded representation error is needed to control the sparsity, i.e., $\|\boldsymbol{\alpha}_i\|_0$. In the image denoising method using sparse coding [25], $\varepsilon$ is related to noise levels. In other words, if the noise level is low, $\varepsilon$ is set with a smaller value to preserve details of the reconstructed patch during the sparse coding. On the contrary, if the inserted noise is serious, $\varepsilon$ is set with a higher value to loosely reconstruct the input patch. Similarly, we can guess that $\varepsilon$ is related to the amount of the rain structures. In the case of rain removal, the amount of the rain structures can be defined as the average value of the absolute of the spatial gradients that are calculated from the rain regions, i.e., $s_i \leq TH_s$. This averaged gradient value can be mapped to the manually tuned bounded representation error ($\varepsilon$) via a linear function. Certainly, other fitting methods (e.g. regression) can be used. In our experiment, the fitting function is given by

$$\varepsilon = 90.7441 \cdot \left( f_{avg} \left( |\nabla_x \mathbf{x}| + |\nabla_y \mathbf{x}| \right) - 0.1107 \right) + 3 \tag{12}$$

where $\nabla_x \mathbf{x}$ and $\nabla_y \mathbf{x}$ indicates the column vectors filled with vertical and horizontal gradients, respectively. If the input image is normalized to [0-1], $\varepsilon$ should be scaled by 1/255. In (12), if the absolute gradient average is less than 0.1107 (i.e., minimum amount of rain structures), $\varepsilon$ is set by 3. 'Prewitt' edge operator was used to calculate the gradients.

## 4. Additional use of classifiers for final shrinkage map design

Rain regions predicted with a trained classifier, for examples, support vector machine (SVM) or deep convolutional neutral network (DCNN) can be additionally used to generate the final shrinkage map. In the proposed method, masked rain images are used to learn the rain dictionary. Therefore, rain features can be extracted via HOG descriptor or convolutional neutral network (DCNN) from the masked rain images. Similarly, from natural images without rain structures, non-rain features can also be extracted. Then, SVM or DCNN can be trained with the two types of rain and non-rain feature sets. Next, given an input rain image, rain and non-rain regions with different binary labels are predicted with trained classifiers (SVM or DCNN) based on patch unit, and then the predicted binary 'label' map is averaged with the final shrinkage map.

- Matlab functions ('extractHOGFeatures' and 'fitclinear') can be used to collect HOG features from images and train the support vector machine, respectively.

- MatConvNet (http://www.vlfeat.org/matconvnet/) can be used to learn the CNN.